\documentclass[twoside,11pt]{article}
\usepackage{jair, theapa, rawfonts}
\usepackage{dblfloatfix}  
\usepackage{booktabs}
\usepackage[linesnumbered,ruled,vlined]{algorithm2e}
\usepackage{graphicx}
\usepackage{subfig}
\usepackage[utf8]{inputenc}
\usepackage[noend]{algpseudocode}
\usepackage{mathtools}
\usepackage{amsmath}
\usepackage{amsfonts}
\usepackage{array}
\usepackage[dvipsnames]{xcolor}
\usepackage{tikz}
\usepackage{float}
\def\checkmark{\tikz\fill[scale=0.4](0,.35) -- (.25,0) -- (1,.7) -- (.25,.15) -- cycle;}
\def\bc{I}
\usetikzlibrary{trees,arrows,automata,positioning}
\graphicspath{ {./images/} }
\SetKwComment{Comment}{$\triangleright$\ \tiny}{}

\SetCommentSty{mycommfont}
\def\mainSense{\textit{mainSense}}
\def\otherSenses{\textit{otherSenses}}
\def\singleWordLabels{\textit{singleWordLabels}}
\def\splitOtherSense{\textit{splitOtherSense}}
\def\splitMainSense{\textit{splitMainSense}}
\def\word{\textit{word}}
\def\len{\textit{len}}
\def\sense{\textit{sense}}
\def\lemmaSynsets{\textit{lemmaSynsets}}
\def\synsets{\textit{synsets}}
\def\nextLevelSynsets{\textit{nextLevelSynsets}}
\def\level{\textit{level}}
\def\synset{\textit{synset}}
\def\edges{\textit{edges}}
\def\edge{\textit{edge}}

\jairheading{}{2021}{}{00/00}{00/00}
\ShortHeadings{Playing Codenames with Language Graphs and Word Embeddings}
{Koyyalagunta, Sun, Draelos, \& Rudin}
\firstpageno{1}

\begin{document}

\title{Playing Codenames with Language Graphs and Word Embeddings}

\author{\name Divya Koyyalagunta* \email divyakoyy@gmail.com \\
       \name Anna Sun* \email anna.sun@alumni.duke.edu \\
       \name Rachel Lea Draelos \email rlb61@duke.edu  \\
       \name Cynthia Rudin \email cynthia@cs.duke.edu \\
        \addr Department of Computer Science\\ Duke University \\
       Durham, NC 27708, USA \\
       *contributed equally}

\maketitle

\begin{abstract}
Although board games and video games have been studied for decades in artificial intelligence research, challenging word games remain relatively unexplored. 
Word games are not as constrained as games like chess or poker. Instead, word game strategy is defined by the players' understanding of the way words relate to each other. 
The word game Codenames provides a unique opportunity to investigate common sense understanding of relationships between words, an important open challenge. We propose an algorithm that can generate Codenames clues from the language graph BabelNet or from any of several embedding methods -- word2vec, GloVe, fastText or BERT. We introduce a new scoring function that measures the quality of clues, and we propose a weighting term called DETECT that incorporates dictionary-based word representations and document frequency to improve clue selection. We develop BabelNet-Word Selection Framework (BabelNet-WSF) to improve BabelNet clue quality and overcome the computational barriers that previously prevented leveraging language graphs for Codenames. Extensive experiments with human evaluators demonstrate that our proposed innovations yield state-of-the-art performance, with up to 102.8\% improvement in precision@2 in some cases. Overall, this work advances the formal study of word games and approaches for common sense language understanding. 
\end{abstract}

\section{Introduction}
\label{Introduction}

\begin{figure*}[!ht]
    \centering
    \includegraphics[scale=0.4]{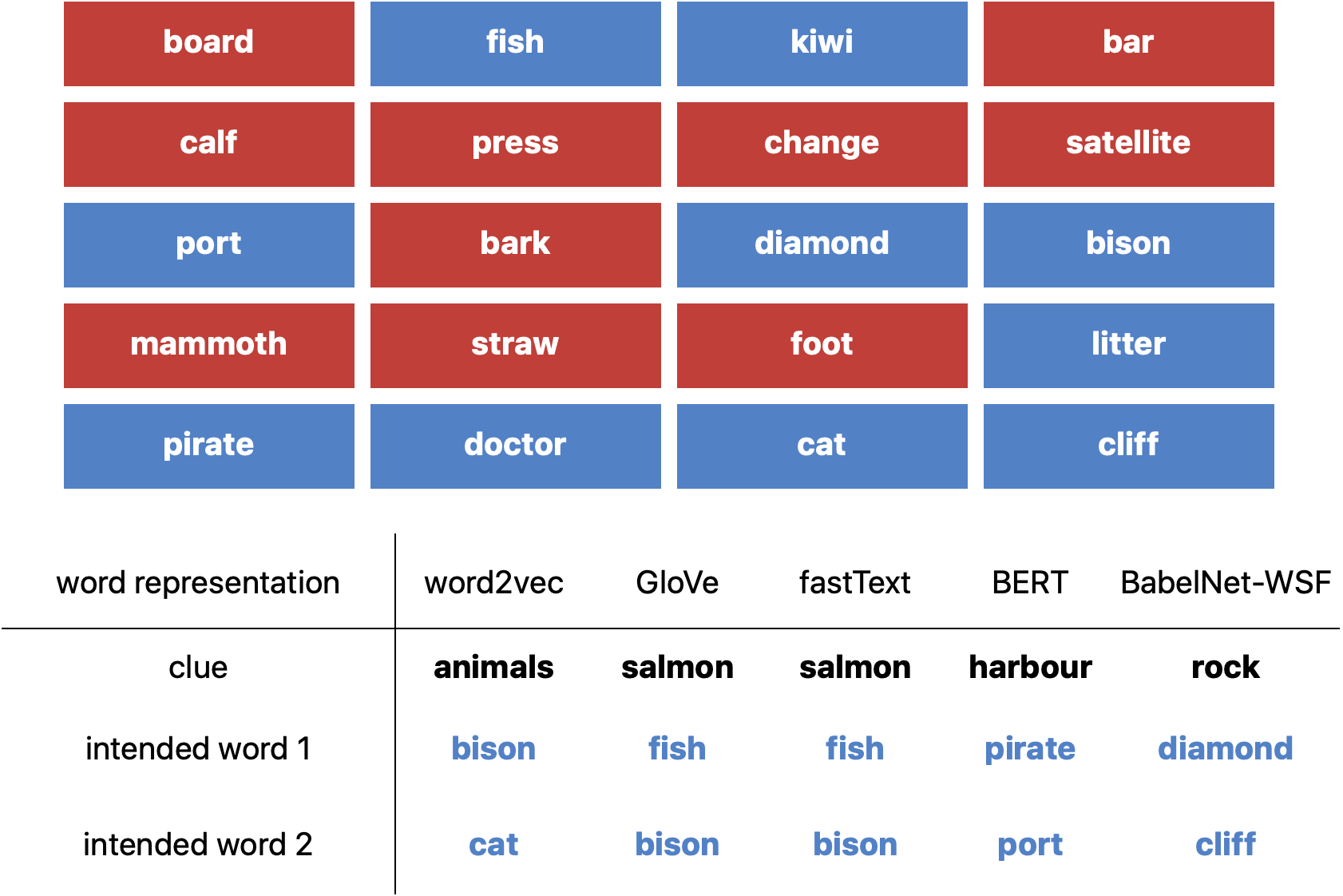}
    \caption{Example of a simplified version of a Codenames board. Blue words belong to the blue team and red words to the red team. Only the clue-givers can see which words belong to which teams. A clue-giver on the blue team generates clues to induce the blue team guessers to pick blue words. The team that identifies all of their own words first wins. In this figure, the table shows clues chosen for the blue team by applying our DETECT algorithm, using various word representations. ``Animals'' is an example of a clue that is too generic, as it also applies to the red words ``calf'' and ``mammoth.'' ``Salmon'' is a good clue for the intended word ``fish'' but less likely to induce a guesser to choose the intended word ``bison'' -- however, ``pirate'' and ``port'' are the only other related words, which are both blue and therefore will not result in a penalty. ``Harbour'' successfully indicates ``pirate'' and ``port'' without applying to any red words, meaning it is a high-quality clue. Similarly, ``rock'' connects to ``diamond'' and ``cliff'' without any close relationships to red words.}
    \label{fig:board_examples}
\end{figure*}

\begin{figure*}[ht]
\centering
\includegraphics[scale = 0.35]{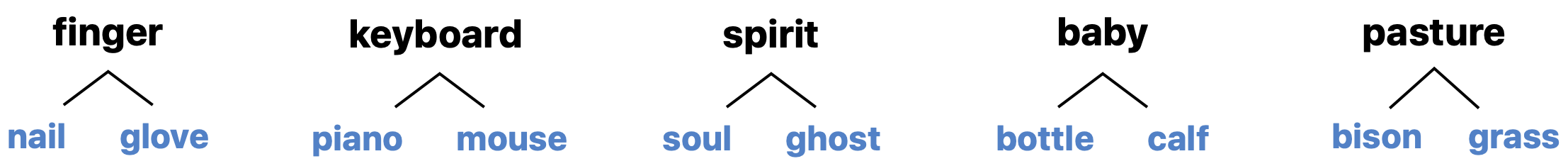}
\caption{Clues chosen for board words by one of our algorithms using various word representations.}
\label{fig:clue_examples}
\end{figure*}

If you wanted to cue the words ``piano'' and ``mouse,'' but not ``bison'' or ``tree'' would you have immediately thought of the clue ``keyboard''? If so, perhaps you are an expert at the game Codenames. This clue, however, was not provided by a human expert -- rather, it was provided by an automated Codenames player that we introduce in this work.

For decades, games have served as a valuable testbed for research in artificial intelligence \cite{yannakakis2018artificial}. Deep neural networks have outperformed human experts in chess \cite{campbell2002deep}, Go \cite{silver2016mastering}, and StarCraft \cite{vinyals2019grandmaster}. In comparison, progress in AI for word games is more limited. The study of word games has immense potential to facilitate deeper insight into how well models represent language, particularly common sense relationships between words.

Codenames is a word game in which a clue-giver must analyze 25 board words and choose a clue word that connects to as many of their own team's words as possible, while avoiding the opposing team's words (\textbf{Figure \ref{fig:board_examples}}). Only the clue-giver knows which words belong to their team, so it is critical that the clue-giver avoid selecting a clue that will cause their team members to guess the opposing team's words. Choosing quality clues requires understanding complex semantic relationships between words, including linguistic relationships such as syno-, anto-, hyper-, hypo- and meronymy, references to history and popular culture, and polysemy (the multiple meanings associated with one word). For example in \textbf{Figure \ref{fig:clue_examples}}, the clue ``keyboard'' connecting the words ``piano'' and ``mouse'' uses polysemy (a keyboard can refer to a musical keyboard or a computer keyboard), hyponymy (a keyboard \textsc{is-a} piano), and context (a computer keyboard is generally accompanied by a mouse). Thus, Codenames is distinct from other natural language processing tasks such as word sense disambiguation (e.g., \shortciteA{iacobacci2016embeddings}) which focuses solely on polysemy, machine translation (e.g., \shortciteA{devlin2014fast}) which focuses on cross-lingual understanding, or part-of-speech tagging (e.g., \shortciteA{collobert2011natural}) which focuses on the grammatical role of words in a sentence. 

Previous work on Codenames has leveraged word embedding models as both clue-givers and guessers, in order to evaluate performance based on how many times a model wins when paired with another model \cite{kim2019cooperation,jaramillo2020word}. This performance evaluation measures how closely the embedding spaces of two methods align, and does not indicate the actual clue quality as judged by a human player. For example, the clue ``duke'' chosen for the board word ``slug'' from word2vec results in a correct guess by a word2vec guesser, but would likely result in a miss by a human guesser. Thus, when one word embedding method judges another, it is possible for the clue-giver embedding method to obtain ``high performance'' in spite of producing nonsensical clues. Instead of using computer simulations for our evaluations, we conduct extensive experiments with Amazon Mechanical Turk to validate the performance of our algorithms through human evaluation.

Previous work has been unable to leverage language graphs for Codenames due to computational barriers. We introduce a new framework, BabelNet-WSF, that improves computational performance by \textit{caching subgraphs} and \textit{introducing constraints on the types of graph traversals that are allowed}. The constraints also improve clue quality by permitting traversals only along paths for which the starting and ending node remain conceptually connected to each other. Additionally, we introduce a method that \textit{extracts semantically relevant single-word clues from BabelNet synsets}, which are groupings of synonymous words associated with a node in the graph. On the whole, BabelNet-WSF enables competitive Codenames performance while remaining broadly applicable to other downstream tasks, including word sense disambiguation (e.g., \citeA{navigli2013semeval}) and semantic relatedness tasks (e.g., \citeA{navigli2012babelrelate}).

In addition to our proposed methods for knowledge graphs, we introduce techniques that improve Codenames performance for both word embedding-based and knowledge-based methods. Most embedding methods rely on a word's context to generate its vector representation. There are two main limitations to this context-based approach for the Codenames task: (1) the resulting embeddings are capable of placing rare words (typically bad clues) close to common words, and (2) important relationships such as meronymy and hypernymy are difficult to capture in embedding methods. In the case of meronymy for example, parts of things (e.g. ``finger'') do not necessarily have the same context as the whole thing (e.g. ``hand''). To address these limitations, \textit{we propose} DETECT, (DocumEnT frEquency $+$ diCT2vec), \textit{a scoring approach that combines document frequency, a weighting term that favors common words over rare words, with an embedding of dictionary definitions}.  Dictionary definitions more effectively capture meronymy, synonymy, and fundamental semantic relationships. For the embedding of dictionary definitions, we use Dict2Vec \cite{tissier2017dict2vec}, though our method can be used with other dictionary-based embeddings. DETECT significantly improves clue quality, with an increase of up to 102.8\% precision@2 above baseline algorithms when evaluated by human players. Furthermore, DETECT leads to universal improvement on Codenames across all four word embedding methods (word2vec, GloVe, fastText, BERT) as well as an improvement on BabelNet-WSF. 

The fact that our methods lead to improvement across all word representation methods is significant, especially in the context of recent work that suggests different word representation methods may be best suited to different sub-tasks. For example, CBOW outperforms GloVe on a Categorization task (clustering words into categories), but GloVe outperforms CBOW on Selectional Preference (determining how typical it is for a noun to be the subject or object of a verb) \shortcite{schnabel2015evaluation}. DETECT is a promising metric to re-weight word similarities in embedding space and in knowledge graphs anywhere that word representations are used, such as comment analysis or recommendation engines.  

This work shows promising results on a difficult word game task that involves many layers of language understanding and human creativity. Because we have focused on human evaluation, we identified problem areas in which word representations fail to perform well, and propose solutions that improve performance dramatically. Overall, our proposed methods advance the formal study of word games and the evaluation of word embeddings and language graphs in their ability to represent common sense language understanding.

\section{Related Work}

The closest related work to ours is that of \citeA{kim2019cooperation}, who proposed an approach for Codenames clue-giving that relies on word embeddings to select clues that are related to board words. They evaluated the performance of word2vec and GloVe Codenames clue-giver bots by pairing them with word2vec and GloVe guesser bots. Although this evaluation approach is easy to run repeatedly over many trials, as it is purely simulation-based, the evaluation is limited to how well the clue-givers and guessers ``cooperate'' with one another. ``Cooperation'' measures how well GloVe and word2vec embedding representations of words are aligned on similarity or dissimilarity of given words. To be more explicit, two methods with different embeddings ``cooperate well'' if word embeddings that are relatively close based on the clue-giver's embeddings are also close based on the guesser's embeddings, and vice versa. As a result, it is clear that perfect performance (100\% win percentage) comes from pairing a clue-giver and a guesser who share the same embedding method. However, this ``cooperation'' metric does not evaluate whether a clue given by a clue-giver is actually a good clue -- that is, a clue that would make sense to a human.

To address this limitation, in this paper, we evaluate Codenames clue-givers based on human performance on the task of guessing correct words given a clue generated by an algorithm.

\citeA{kim2019cooperation} also explored the use of knowledge graphs for Codenames clue-giving, but ultimately did not consider knowledge-graph-based clue-givers in their final evaluation due to poor qualitative performance and computational expense. In contrast, we propose a method for an interpretable knowledge-graph-based clue-giver that performs competitively with embedding-based approaches. The knowledge-graph-based method has a clear advantage in interpretability over the embedding-based approaches.

In an extension of \citeA{kim2019cooperation},  \citeA{jaramillo2020word} compared the baseline word2vec + GloVe word representations with versions using TF-IDF values, classes from a naive Bayes classifier, or nearest neighbors of a GPT-2 Transformer representation of the concatenated board words. Similar to \citeA{kim2019cooperation}, they evaluated their methods primarily by pairing clue-giver bots with guesser bots. They included an initial human evaluation, where 10 games were played for both the baseline (word2vec + GloVe) and the Transformer representations as clue-giver and guesser, but the human evaluation is limited to only 40 games. Again, since evaluations from bots may not represent human judgments, our human evaluation is more realistic and extensive, conducted through Amazon Mechanical Turk with 1,440 total samples.

Another method \cite{zunjani2019towards} proposes a formalization of the Codenames task using a knowledge graph, but does not provide an implementation of their proposed recursive traversal algorithm. We found that recursive traversal does not scale to the computation required to run repeated evaluations of Codenames - for each blue word, we must find each associated word $w$ in the knowledge base that have $Association_{w,b}$ greater than some threshold $t$, and repeat this process every trial. In BabelNet, because each word may be connected to tens or hundreds of other words, this becomes unscalable when traversing more than one or two levels of edges. We propose an approach that scales significantly better than naive recursive traversal by limiting paths through the graph to those that yield high-quality clues.

\shortciteA{shen2018comparing} also propose a simpler version of the Codenames task with human evaluation. The experimental setup focuses on comparing different semantic association metrics, including a knowledge-graph based metric. Their task differs from ours in two key ways. First, each of their trials considers three candidate clues drawn from a vocabulary of 100 words, whereas we consider candidate clues drawn from the larger vocabulary of all English words. Second, their usage of ConceptNet is different from our usage of BabelNet because they use vector representations derived from an ensemble of word2vec, GloVe, and ConceptNet using retrofitting \cite{speer2017conceptnet} whereas we leverage the graph structure of BabelNet.

\begin{figure}
    \centering
    \includegraphics[scale=0.23]{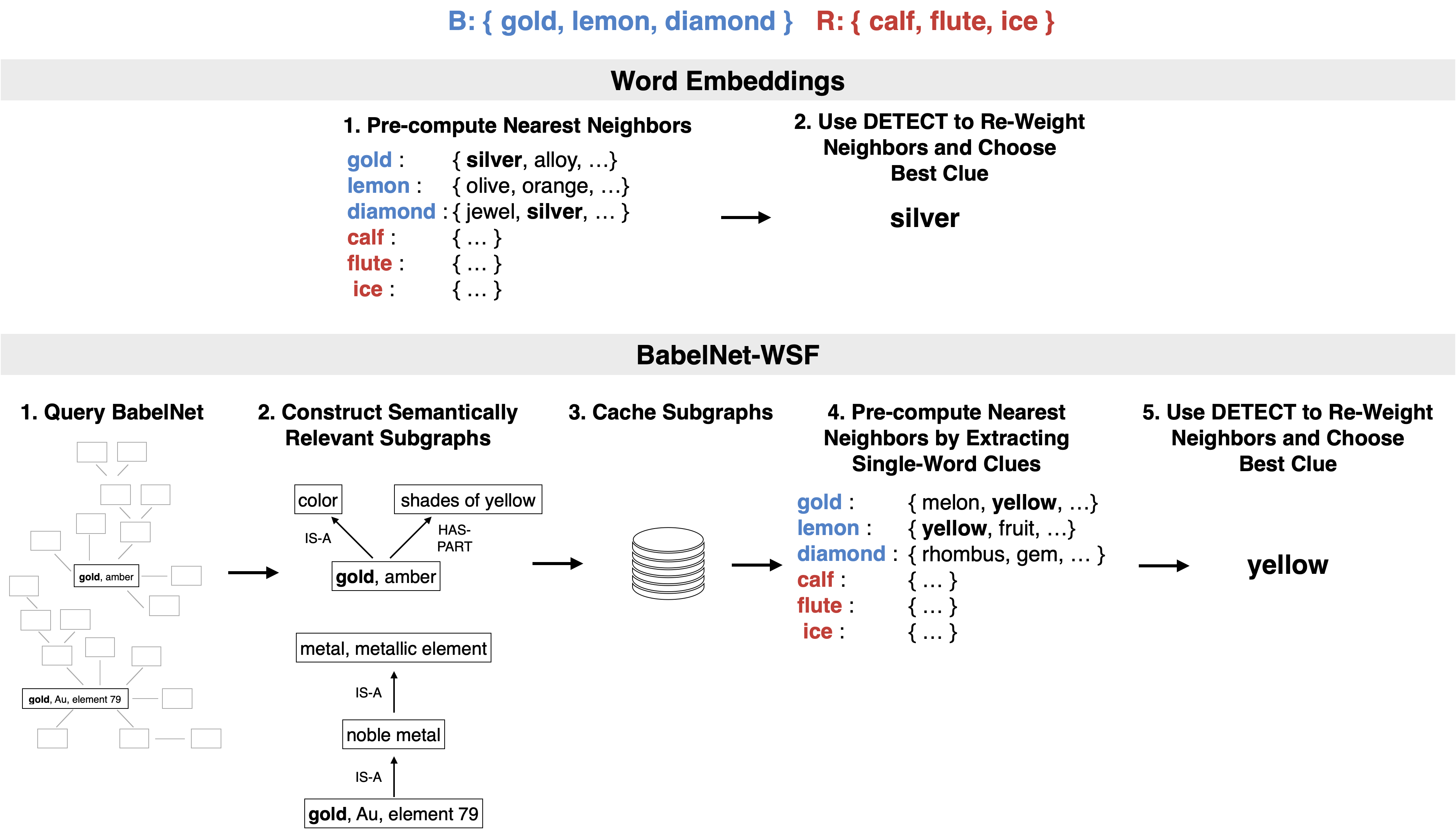}
    \caption{An overview of our proposed methods for improving Codenames performance. For word embedding approaches, this involves pre-computing nearest neighbors for all board words (e.g., gold, lemon, diamond, calf, flute and ice as shown above). To pre-compute nearest neighbors using BabelNet-WSF, we query BabelNet for every board word, and then construct the semantically relevant subgraph connected to that board word (described in \textbf{Section \ref{sec:babelnetfilternn}}). We then cache the subgraphs for later use (see \textbf{Section \ref{sec:constructing_subgraph}}), and extract single-word clues from BabelNet synsets (see \textbf{Section \ref{sec:single_word_clues}}). In \textbf{Section \ref{sec:bertembeddings}}, we describe how we get single word embeddings from a contextual embedding based method such as BERT. Once candidate clues are produced by either a word embedding or BabelNet-WSF, we use the ClueGiver algorithm described in \textbf{Section \ref{sec:ClueGiver}} and apply our DETECT weighting term to choose the best clue for that board (see \textbf{Section \ref{sec:DETECT}}).}
    \label{fig:paperoverview}
\end{figure}

\subsection{Other language games} 
\shortciteA{ashktorab2021effects} propose three different AI approaches for a word game similar to Taboo, including a supervised model trained on Taboo card words, a reinforcement learning model, and count-based model using a word evocation dataset. Their task, in which an agent gives clues until a user guesses the secret word, is different from Codenames. However, the datasets used in their models present interesting directions for future work in the Codenames task.
\shortciteA{ashktorab2021effects} propose three different AI approaches for a word game similar to Taboo, including a supervised model trained on Taboo card words, a reinforcement learning model, and count-based model using the Small World of Words \shortcite{de2019small}, a word evocation dataset. Their task, in which an agent gives clues until a user guesses the secret word, is different from Codenames. However, the word evocation dataset used in their count-based model could be leveraged for the Codenames task since it represents word relatedness, and presents interesting directions for future work.
Another related task is the Taboo Challenge competition \cite{rovatsos2018taboo}, where AI systems must guess a city based on clues crowdsourced from humans. In this task, the AI system  acts as the guesser, rather than the clue giver.

Other work in language games that are similar to Codenames include the Text-Based Adventure competition \shortcite{atkinson2019text}, which evaluates agents in text-based adventure games, \citeA{xu2010inference}, which models a task in which a speaker gives a one-word clue to a listener with the goal of guessing a target word, and \citeA{thawani2019swow}, which proposes a task to evaluate embeddings based on human associations.

\subsection{Quantifying Semantic Relatedness from WordNet}
There is a body of previous work that proposes methods to quantify semantic relatedness in a knowledge graph beyond simply counting the number of edges between two nodes. \citeA{hirst1998lexical} developed a ranking of relationships into three groups, extra-strong, strong, and medium-strong, where only medium-strong includes a numeric score based on path length and path type; in our work we use numeric scores for all word comparisons and do not separate into 3 categories. \citeA{budanitsky2006evaluating} evaluated different techniques for the quantification of semantic relatedness for WordNet, but most techniques used noun-only versions of WordNet, whereas we consider all parts of speech including nouns, verbs, and adjectives. Furthermore, 4 of their 5 techniques restrict to hyponym relationships whereas we consider multiple relationship types. The proposed PageRank technique of \shortciteA{agirre2009study} could be applied to BabelNet and is an interesting direction for future work.

\section{Methods}

Codenames is a word-based, multiplayer board game illustrated in \textbf{Figure \ref{fig:board_examples}}. The board consists of a total of $2N$ words, divided equally into the blue team's words $B=\{b_n\}_{n=1}^N$ and the red team's words $R=\{r_n\}_{n=1}^{N}$. Only the clue-giver knows which words are assigned to which teams. A clue-giver provides a clue, and a guesser on their team selects a subset of board words related to that clue. The team that guesses all of their own words first wins. 

In this paper we focus on the task of clue-giving. The clue-giving task is the most interesting, since the clue-giver needs to search the space of all possible words to identify suitable clues, and then rank the clues to select the best one. A blue clue-giver must generate a clue $c$ that is conceptually closest to one subset (here, a pair) of blue words from the set of all possible blue word pairs $\bc \in  B^2$. The clue $c$ must also be sufficiently far from all red words $R$. Note that in the original Codenames game, a clue-giver can produce a clue corresponding to an arbitrary number of blue words from $\bc \in  B^m$ for any $m \leq N$, but for more consistent evaluation we calculate performance on the clue-giving task considering $m=2$ only. Our proposed innovations apply for arbitrary $m$.

\textbf{Figure \ref{fig:paperoverview}} provides an overview of our proposed methods. In \textbf{Section \ref{sec:ClueGiver}}, we describe the baseline algorithm for choosing and ranking clues. \textbf{Section \ref{sec:babelnetgeneral}} details BabelNet-WSF, our method for querying the very large BabelNet graph and multiple techniques to improve the quality of the returned clues. In \textbf{Section \ref{sec:DETECT}} we propose the DETECT algorithm which leverages document frequency and dictionary embeddings to universally improve Codenames performance across embedding and graph-based clue-givers. Finally, \textbf{Section \ref{sec:humaneval}} explains the extensive human evaluation experiments conducted through Amazon Mechanical Turk.

All of our code for these proposed methods is available for public use.\footnote{https://github.com/divyakoyy/codenames}

\subsection{ClueGiver: a baseline algorithm that produces clues} \label{sec:ClueGiver}

We propose ClueGiver, an algorithm that gives clues on the basis of a measurement of word similarity $s(w_1,w_2)$. 

First, we define $s(w_1,w_2)$ for word-embedding-based and knowledge-graph-based approaches. 
For word embeddings, the word similarity $s(w_1,w_2)$ is defined as 1 minus the cosine distance between the word embeddings, \textit{i.e.} $1 - \cos(f(w_1), f(w_2))$, where $f$ is the embedding function. When measuring word similarity with a knowledge graph method such as BabelNet-WSF, the word similarity $s(w_1,w_2)$ is defined as the inverse of the number of edges along the path between the graph node for word $w_1$ and the graph node for word $w_2$, \textit{i.e.} $ \frac{1}{h(w_1, w_2) + 1}$, where $h$ is a function that gives the number of edges along the shortest path between $w_1$ and $w_2$.

ClueGiver has two steps. To choose a clue for the blue team, we first calculate the $T$ nearest neighbors of each blue word in $B=\{b_n\}_{n=1}^N$. We go through each subset $I \in B^m$ and add the union of the nearest neighbors for every blue word $b \in \bc$ to a set of candidate clues $\tilde{C}=\{\tilde{c}_{tn}\}_{t=1..T;n=1..N}$. The subset $\bc$ represents the set of intended words that are meant to match the candidate clue. Every subset of $B$ of size $m$ is therefore considered as a candidate set of intended words. Next, we score each candidate clue $\tilde{c}$ using the following scoring function $g(\cdot)$ which produces a large positive value for good clues and a lower value for bad clues (and thus should be maximized): 

\begin{equation}
\begin{aligned}
g(\tilde{c}, \bc) = \quad & \lambda_B \left( \sum_{b \in \bc}s( \tilde{c} , b) \right) -\lambda_R \left( \max_{r \in R} s( \tilde{c} , r)\right) 
\end{aligned}
\label{eqn:our_scoring_fxn}
\end{equation}

The final chosen clue $c$ is the candidate clue with the highest score. A candidate clue $\tilde{c}$ will have a high score if it is closest to the subset $\bc$ (thus making $\lambda_B \big{(} \sum_{b \in \bc}s( \tilde{c} , b) \big{)}$ as large as possible), while remaining as far away as possible from all the red words. The expression $\max_{r \in R} s( \tilde{c}, r)$ means that we calculate the highest similarity between the red words and our candidate clue $\tilde{c}$, which corresponds to the red word closest to $\tilde{c}$. If $\tilde{c}$ is a good clue for the blue team, then even this closest red word is far away from the candidate clue, with a small positive value of $s(\tilde{c}, r)$. In the case that there is no overlap between the nearest neighbors of the blue words, the algorithm will choose a clue for one word. This happened extremely rarely when computing 500 nearest neighbors for each board word.
The choices of $\lambda_B$ and $\lambda_R$ determine the relative importance of each part of the scoring function, \textit{i.e.}, whether we should prioritize clues that are close to blue words or prioritize clues that are far from red words. We found $\lambda_B = 1$ and $\lambda_R = 0.5$ to be effective values empirically across all word representations. None of the human experiments used the same data that was used to set these parameters. 
The Codenames boards used to tune the parameters were randomly sampled from the total of $208 \choose 20$ = 3.68e+27 Codenames boards (208 being the possible board words, and 20 being the board size). The Codenames boards used for human evaluation on AMT were different boards, randomly sampled from all possible boards with each having probability $\frac{1}{3.68e+27}$.

\textbf{Comparison to the scoring function of \citeA{kim2019cooperation}} In our experiments, we compare our scoring function $g(\cdot)$ with the following scoring function proposed by \citeA{kim2019cooperation}. The term $\lambda_T$ is configurable to limit how aggressive the clue-giver is:

\begin{equation}
    \begin{aligned}
    g_{kim}(\tilde{c}, \bc) = 
    \begin{cases}
    \min_{b \in \bc}s( \tilde{c}, b), & \text{if } \min_{b \in \bc}s( \tilde{c} , b) > \lambda_T \text{ and } \min_{b \in \bc}s( \tilde{c} , b) > \max_{r \in R}s( \tilde{c} , r)\\
    0,              & \text{otherwise}
    \end{cases} 
    \end{aligned}
    \label{eqn:kim_scoring_fxn}
\end{equation}

The main difference between $g_{kim}$ and our proposed scoring function $g$ is that our scoring function $g$ incorporates a penalty to the score based on the similarity of the closest red word, while $g_{kim}$ enforces the constraint that the similarity between the clue and the furthest blue word must be greater than the similarity to the closest red word. In addition, $g_{kim}$ enforces that the similarity between the furthest blue word and the clue is greater than the threshold $\lambda_T$. Notably, $g_{kim}$ would give equal scores (assuming that blue word distances are equal) to clue words that do not violate the constraint, even if one was much closer to a red word than the other, whereas $g$ applies a soft penalization to red words.

\subsection{BabelNet-WSF: Solving Codenames clue-giving with the BabelNet knowledge graph} \label{sec:babelnetgeneral}

As discussed previously, knowledge graphs have not been successfully used to play Codenames in prior work.
In this section we propose BabelNet-WSF, which includes three innovations to enable high-performance use of the BabelNet knowledge graph for the Codenames clue-giving task, including (1) a method for constructing a nearest neighbor BabelNet subgraph relevant to a particular Codenames board, (2) constraints that filter the nearest neighbors to improve candidate clue quality, and (3) an approach to select a good single word clue from the set of synonymous phrases associated with a particular node.

\subsubsection{Constructing Subgraphs of BabelNet to Identify Nearest Neighbors} \label{sec:constructing_subgraph}

In order to choose a clue using the scoring function defined in the previous section, it is necessary to identify the nearest neighbors of the board words. When solving Codenames with a knowledge graph like BabelNet, the graph connectivity defines the nearest neighbors. We define the ``nearest neighbors'' of an origin node as all nodes within 3 edges of the origin node. The challenge arises because the full BabelNet 4.0.1 graph is 29 gigabytes, so recursively traversing it to identify nearest neighbors is computationally prohibitive as noted by \citeA{kim2019cooperation}.

We propose three steps to enable fast nearest neighbors identification from BabelNet for a particular Codenames board. First, we restrict the relationship edges that are added to the subgraph. Every edge in the BabelNet graph indicates a particular type of relationship (\textit{e.g.} \textsc{is-a}). For each node in the graph representing a board word, we obtain all outgoing edges for the first edge, but for edges beyond that, we only recursively traverse edges that are in the \textsc{hypernym} relationship group, as discussed in \textbf{Section \ref{sec:babelnetfilternn}}.

Second, we exclude edges that were automatically generated, because we found these to be of poor quality for the Codenames task. Finally, we cache the nearest neighbor results for each board word, in order to reuse these results for future boards. This caching step is important due to a daily limit in querying the BabelNet API. Appendix \textbf{Algorithm \ref{algo:3}} details the overall process for obtaining nearest neighbors from BabelNet.

\subsubsection{Filtering Nearest Neighbors for BabelNet} \label{sec:babelnetfilternn}

Once a nearest neighbor graph is constructed, we apply two filtering steps on the edges in order to retain only the highest-quality nearest neighbors, and thereby improve the selected clues.

\textit{Hypernym edge constraint beyond the first edge}. The first filtering step is to only allow hypernym relationships (\textsc{is-a} or \textsc{subclass-of}) beyond the first edge. The motivation behind this constraint is that traversing the graph using randomly chosen relationship types leads to ``nearest neighbors'' which are not very related to the origin node. However, using only hypernym relationship types is too restrictive, since exploiting different kinds of relationships is critical in order to perform well at Codenames clue-giving. We found that allowing the first edge to be any relationship type, but restricting all subsequent edges to be a hypernym relationship type maintained diversity of clues while usually preserving conceptual relatedness of the origin node and (possibly distant) neighbor node. Table \ref{table:hypernymconstraint} illustrates the improvement in nearest neighbor quality that this constraint yields.

\begin{table}[ht]
\centering
{\fontfamily{phv}\selectfont
\def\arraystretch{1.75}%
 \begin{tabular}{|c|c|} 
 \hline
 passes constraint &    \textbf{\textcolor{Green}{needle} \textcolor{TealBlue}{$\xrightarrow{\textsc{has-part}}$} \textcolor{Green}{point}} \\
 \hline
 fails constraint & \textbf{\textcolor{Green}{needle} \textcolor{TealBlue}{$\xrightarrow{\textsc{has-part}}$} \textcolor{Green}{point} \textcolor{magenta}{$\xrightarrow{\textsc{has-kind}}$} \textcolor{Red}{spearhead}} \\
 \hline\hline
 passes constraint & \textbf{\textcolor{Green}{litter} \textcolor{blue}{\textcolor{blue}{$\xrightarrow{\textsc{is-a}}$}} \textcolor{Green}{trash} \textcolor{blue}{\textcolor{blue}{$\xrightarrow{\textsc{is-a}}$}} \textcolor{Green}{waste}} \\
 \hline
 fails constraint & \textbf{\textcolor{Green}{litter} \textcolor{blue}{\textcolor{blue}{$\xrightarrow{\textsc{is-a}}$}} \textcolor{Green}{trash} \textcolor{magenta}{$\xrightarrow{\textsc{has-kind}}$} \textcolor{Red}{scrap metal}} \\ 
 \hline\hline
 passes constraint & \textbf{\textcolor{Green}{mouse} \textcolor{blue}{\textcolor{blue}{$\xrightarrow{\textsc{is-a}}$}} \textcolor{Green}{pointing device} \textcolor{blue}{\textcolor{blue}{$\xrightarrow{\textsc{is-a}}$}} \textcolor{Green}{input device}} \\
 \hline
 fails constraint & \textbf{\textcolor{Green}{mouse} \textcolor{blue}{\textcolor{blue}{$\xrightarrow{\textsc{is-a}}$}} \textcolor{Green}{pointing device} \textcolor{magenta}{$\xrightarrow{\textsc{has-kind}}$} \textcolor{Red}{light gun}} \\ 
 \hline
\end{tabular}
}
\caption{Examples of hypernym edge constraint beyond the first edge. Hypernym relationships are \textsc{is-a} or \textsc{subclass-of}. Words in red failed the constraint and are filtered out of the graph.}
\label{table:hypernymconstraint}
\end{table}

\textit{Same-type edge constraint}. The second constraint restricts all edges after the first edge to be the same relationship type. In combination with the hypernym constraint, that means a traversal after the first edge of \textsc{is-a}/\textsc{is-a}/\textsc{is-a} is allowed, a traversal after the first edge of \textsc{subclass-of}/\textsc{subclass-of}/\textsc{subclass-of} is allowed, but any traversal after the first edge randomly combining \textsc{is-a} and \textsc{subclass-of} is forbidden. We found that this additional stringency improved the relevance of retrieved nearest neighbors. Examples are shown in Table \ref{table:sametype_edgeconstraint}.

\begin{table}[ht]
\def\arraystretch{1.75}%
\centering
{\fontfamily{phv}\selectfont
\resizebox{\textwidth}{!}{%
 \begin{tabular}{|c|c|} 
 \hline
 passes constraint    &  \textbf{\textcolor{Green}{litter} \textcolor{blue}{$\xrightarrow{\textsc{is-a}}$} \textcolor{Green}{animal group} \textcolor{blue}{$\xrightarrow{\textsc{is-a}}$} \textcolor{Green}{biological group} \textcolor{blue}{$\xrightarrow{\textsc{is-a}}$} \textcolor{Green}{group}}\\
 \hline
 passes constraint &  \textbf{\textcolor{Green}{litter} \textcolor{blue}{$\xrightarrow{\textsc{is-a}}$} \textcolor{Green}{animal group} \textcolor{Purple}{$\xrightarrow{\textsc{subclass-of}}$} \textcolor{Green}{fauna}}  \\
 \hline
 fails constraint & \textbf{\textcolor{Green}{litter} \textcolor{blue}{$\xrightarrow{\textsc{is-a}}$} \textcolor{Green}{animal group} \textcolor{Purple}{$\xrightarrow{\textsc{subclass-of}}$} \textcolor{Green}{fauna} \textcolor{blue}{$\xrightarrow{\textsc{is-a}}$} \textcolor{Red}{aggregation}} \\ 
 \hline\hline
 passes constraint & \textbf{\textcolor{Green}{moon} \textcolor{TealBlue}{$\xrightarrow{\textsc{gloss-related}}$} \textcolor{Green}{planet} \textcolor{blue}{$\xrightarrow{\textsc{is-a}}$} \textcolor{Green}{celestial body} \textcolor{blue}{$\xrightarrow{\textsc{is-a}}$} \textcolor{Green}{natural object}} \\ 
 \hline
 fails constraint & \textbf{\textcolor{Green}{moon} \textcolor{TealBlue}{$\xrightarrow{\textsc{gloss-related}}$} \textcolor{Green}{planet} \textcolor{Purple}{$\xrightarrow{\textsc{subclass-of}}$} \textcolor{Green}{planemo} \textcolor{blue}{$\xrightarrow{\textsc{is-a}}$} \textcolor{Red}{object}} \\ 
 \hline\hline
 passes constraint & \textbf{\textcolor{Green}{figure} \textcolor{TealBlue}{$\xrightarrow{\textsc{gloss-related}}$} \textcolor{Green}{diagram} \textcolor{blue}{$\xrightarrow{\textsc{is-a}}$} \textcolor{Green}{drawing} \textcolor{blue}{$\xrightarrow{\textsc{is-a}}$} \textcolor{Green}{representation}} \\ 
 \hline
 fails constraint & \textbf{\textcolor{Green}{figure} \textcolor{TealBlue}{$\xrightarrow{\textsc{gloss-related}}$} \textcolor{Green}{diagram} \textcolor{Purple}{$\xrightarrow{\textsc{subclass-of}}$} \textcolor{Green}{graphics} \textcolor{blue}{$\xrightarrow{\textsc{is-a}}$} \textcolor{Red}{visual communication}} \\ 
 \hline
\end{tabular}}
}
\caption{Examples of same-type edge constraint on edges after the first edge.}
\label{table:sametype_edgeconstraint}
\end{table}

\subsubsection{Selecting a good single-word clue from a multi-word synset} \label{sec:single_word_clues}

Once the nearest neighbors have been filtered, a final clue must be selected. In the Codenames task, the clue must consist only of a single word. However, each node of the BabelNet graph is a synset, which is a group of related concepts organized as a ``main sense label'' with ``other sense labels.'' To add to the complexity, each label can be one or more words. For example from Table \ref{table:multi_word_synsets}, the synset with the definition ``material used to provide a bed for animals'' has a main sense label of ``bedding,'' and other sense labels  ``litter'' and ``bedding material.'' The synset with the main sense label ``creative work'' has other sense labels ``artwork,'' ``work,'' and ``work of art.'' It is not immediately obvious how to extract a good-quality single word clue from a synset.

\begin{table}[ht]
\centering
\begin{tabular}{m{10em} m{15em} m{10em} } 
\toprule
Main sense label & Definition & Other labels \\ [0.5ex] 
\midrule
bedding & Material used to provide a bed for animals & litter, bedding material \\ 
bedding & Coverings that are used on a bed & bedclothes, bed clothing \\
stringed instrument & A musical instrument in which taut strings provide the source of sound & string instrument, chordophone \\
creative work & A creative work is a manifestation of creative effort including fine artwork, writing, filmmaking, and musical composition & creative work, artwork, work, work of art \\
\bottomrule
\end{tabular}
\caption{Examples of synsets with multi-word labels}
\label{table:multi_word_synsets}
\end{table}

Selecting a single word at random from a synset is not effective. For example ``work'' on its own is not an ideal choice for ``creative work'' and ``material'' is not an ideal choice for ``bedding material.'' We develop a scoring system to select the best possible single word clue from a synset. First, we keep only single words that belong to the intersection of the nearest neighbors of two board words. Taking a simplified example from Figure \ref{fig:paperoverview} (bottom), for the board words ``gold,'' having neighbor synset with labels \{``shades of yellow,'' ``variations of yellow''\}, and ``lemon,'' having neighbor synset with labels \{``yellow,'' ``yellowness,'' ``color yellow''\}, the word ``yellow'' is kept.

We also apply weights (which are parameters of our method) on each of the words comprising the synset labels. The weights on these words are denoted by $w_1$, $w_2$, $w_3$, and $w_4$ where $w_1 \leq w_2 \leq w_3 \leq w_4$, and they are assigned based on the type of synset label and whether the label has multiple words, as shown in \textbf{Table \ref{table:1}}. The number of edges $h( \tilde{c}, w)$ is multiplied by the weight; thus a label type with a lower weight, such as main-sense single word, is desirable. Empirically, we found that the values $w_1 = 1, w_2 = 1.1, w_3 = 1.1, w_4 = 1.2$ to be effective.
Appendix \textbf{Algorithm \ref{algo:2}} details the process for obtaining single-word clues and their corresponding weights.

\begin{table}[ht]
\centering
 \begin{tabular}{c c c} 
 \toprule
 label type & single- or multi- word & weight \\ [0.5ex] 
 \midrule
 main sense & single & $w_1$ \\ 
 main sense & multi & $w_2$ \\
 other sense & single & $w_3$ \\
 other sense & multi & $w_4$ \\
 \bottomrule
\end{tabular}
\caption{Weights applied to labels taken from BabelNet depend on whether the label is a main sense or other sense label (this distinction is provided by BabelNet), and whether that label is composed of one or more words.}
\label{table:1}
\end{table}

\subsubsection{Example BabelNet-WSF Clue}

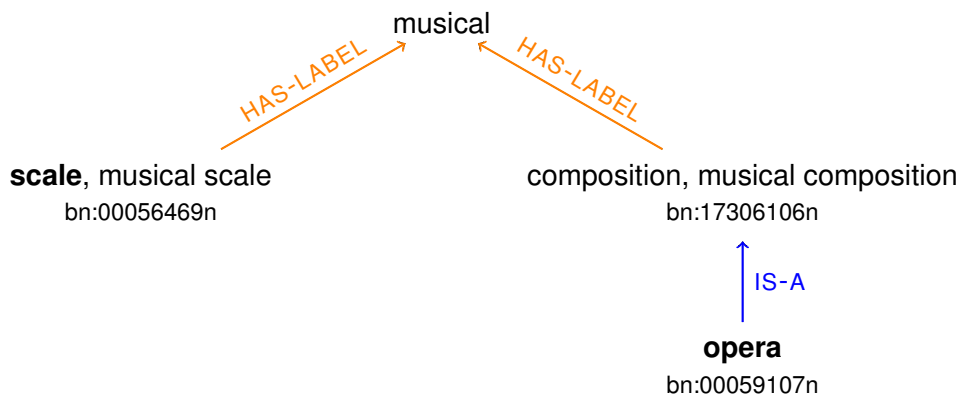
\begin{figure*}[ht]
    \centering
    {\fontfamily{phv}\selectfont
    \begin{tikzpicture}[level distance=2.3cm,
      level 1/.style={sibling distance=8cm},
      level 2/.style={sibling distance=1.5cm},<-]
      \node (A1) {musical}
        child {node (B1) {\begin{tabular}{c} \textbf{scale}, musical scale \\ \footnotesize bn:00056469n \end{tabular} }}
        child {node (B2) {\begin{tabular}{c} composition, musical composition \\ \footnotesize bn:17306106n \end{tabular}}
          child {node (C1) {\begin{tabular}{c} \textbf{opera} \\ \footnotesize bn:00059107n \end{tabular}}}
        };
    \path[<-,draw, orange, thick]
         (A1) edge node [above, rotate=31]  { \textcolor{orange} {\textsc{has-label}}} (B1)
         (A1) edge node [above, rotate=-31] {\textcolor{orange} {\textsc{has-label}}} (B2)
    ;
    \path[<-,draw, blue, thick]
        (B2) edge node[right] {\textcolor{blue} {\textsc{is-a}}} (C1)
    ;
    \end{tikzpicture}
    }
    \caption{Sub-graph of BabelNet showing how the clue ``musical'' was chosen for the board words ``scale'' and ``opera'' using BabelNet-WSF. The \textsc{has-label} edges are not real edges in BabelNet, but rather single word labels that we extract using the single-word clue approach described in Section \ref{sec:single_word_clues}. Synsets from BabelNet are annotated with their ID from BabelNet 4.0.1 as bn:synset-id.}
    \label{fig:babelnet_graph_goodex}
\end{figure*}

The combination of the three aforementioned innovations enables selection of high-quality clues from BabelNet, as exemplified in Figure \ref{fig:babelnet_graph_goodex}. 

\subsection{DETECT for Improving Clue Quality} \label{sec:DETECT}

The previous sections focused on obtaining good clues from BabelNet. In this section, we describe a method called DETECT that improves the quality of clues for both BabelNet-WSF and word embedding-based methods. We originally developed DETECT to address deficiencies in clues produced by word embedding-based methods, and then discovered that DETECT also improved BabelNet-WSF clues. In our experiments, we show results of word embedding and BabelNet-WSF methods with and without DETECT.

We identified three deficiencies in the clues chosen via word embeddings: (1) obscure clues, (2) overly generic clues, and (3) lack of clues exploiting many common sense word relationships. \textbf{Table \ref{table:5}} provides real examples of clues that suffer from deficiencies (1) and (2). Obscure clues are not necessarily ``incorrect'' in the way they connect two words: for example the word ``djedkare'' in \textbf{Table \ref{table:5}} refers to the name of the ruler of Egypt in the 25th century B.C., and therefore correctly connects the words ``egypt'' and ``king.'' However this clue does not reflect the average person's knowledge of the English language and is likely to yield random guesses if presented to a human player. Overly generic clues, in contrast, are more likely to match too many board words. For example ``artifact'' in \textbf{Table \ref{table:5}} connects ``key'' and ``pipe,'' but also matches other red words on the board for that trial, such as ``crown'' and ``racket.''

\begin{table}[ht]
\centering
 \begin{tabular}{ m{5em}  m{15em} } 
 \toprule
 Clue & Intended Words to Match Clue \\ [0.5ex] 
 \midrule
 \textbf{aether} & jupiter, vacuum \\ 
 \textbf{djedkare} & egypt, king \\ 
 \textbf{machine} & key, crown \\ 
 \textbf{artifact} & key, pipe \\ 
 \bottomrule
\end{tabular}
\caption{Examples of obscure and generic clues. Obscure clues include ``aether,'' which in ancient and medieval science is a material that fills the region of the universe above the terrestrial sphere, and  ``djedkare'' which is the name of the ruler of Egypt in the 25th century B.C. Overly generic clues include ``machine'' and ``artifact'' which often apply to many board words.}
\label{table:5}
\end{table}

To address all three issues in clue quality, we added a scoring metric, DETECT, to our scoring function $g(\cdot)$. DETECT includes two parts, a function $FREQ(w)$ that uses document frequency to exclude too-rare and too-common words, and a function $DICT(w_1,w_2)$ that encourages clues relying on common sense word relationships. 

\textit{Leveraging document frequency with $FREQ$.} In order to penalize overly rare as well as overly generic tokens, we leverage the document frequency $f_w$ of a word, which indicates the count of documents in which word $w$ is found in a cleaned subset of Wikipedia. We calculate $FREQ(w)$ as: 

\begin{equation}
\begin{aligned}
FREQ(w) = -\begin{cases} 
      \frac{1}{df_{w}} & \textrm{when } \frac{1}{df_{w}} \geq \alpha  \\
      1 & \textrm{when } \frac{1}{df_{w}} < \alpha.
  \end{cases}
\end{aligned}
\label{eqn:idf}
\end{equation}

\begin{figure*}[t]
    \centering
    \includegraphics[scale=0.33]{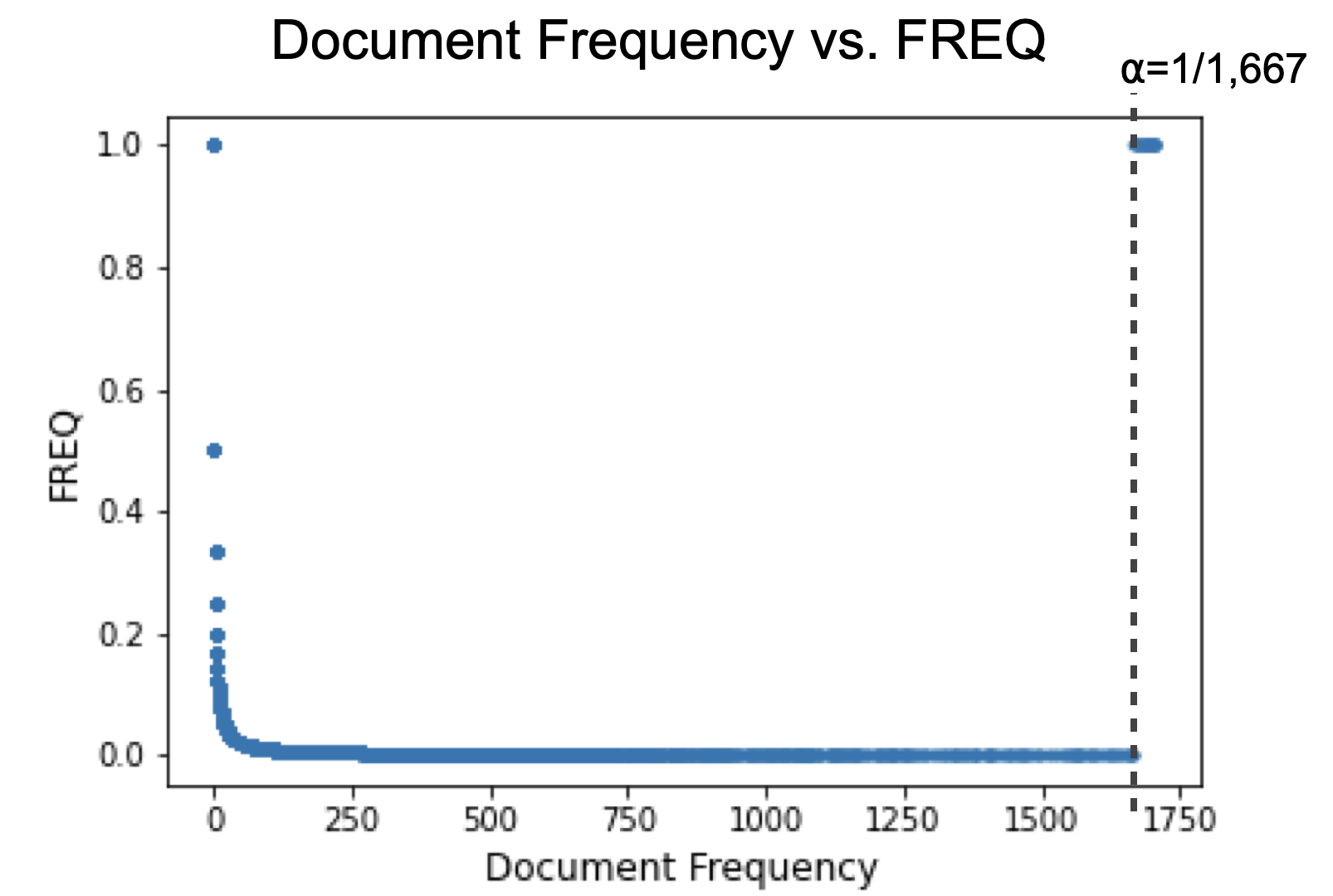}
    \caption{FREQ is a function of document frequency that is used to penalize rare words more and common words less unless the word is so common that it is not useful as a clue word. $\alpha$ was chosen by picking an upper bound document frequency and validating empirically across clue-giving algorithms. Note that $\alpha=1/1,667$ was calculated on a sample of 1,701 cleaned documents from \cite{Mahoney}.}
    \label{fig:doc_freq_vs_freq}
\end{figure*}

$FREQ(w)$ penalizes rare words more and common words less, unless a word is so common that the inverse function of its document frequency is lower than a value $\alpha$, which is an algorithm parameter. $\alpha$ was chosen empirically based on the distribution of document frequencies in a cleaned subset of the Wikipedia corpus as shown in \textbf{Figure \ref{fig:doc_freq_vs_freq}}, together with the clues produced across all algorithms. $\alpha$ represents the upper bound document frequency at which point a word is considered too common to be useful as a clue word. \citeA{jaramillo2020word} used TF-IDF as a standalone baseline method, whereas in this work $FREQ$ is a term which is part of the full scoring function.

In principle, \textbf{Equation \ref{eqn:our_scoring_fxn}} should filter out overly generic clues that match red words via the penalty we apply to red words. However we found that in practice, performance improved by removing common words with the $FREQ$ component of DETECT. This is because of a limitation of BabelNet, in which the number of edges connecting a common word to all of its children is highly variable (e.g. the number of edges between ``object'' and ``apple'' is 4, vs. the number of edges between ``object'' and ``cash'' is 1).

\textit{Leveraging dictionary embeddings with $DICT$.} Since most embedding methods rely on a word's context to generate its vector representation, they do not always encode important relationships such as meronymy and hypernymy. Dict2Vec \cite{tissier2017dict2vec} addresses this issue by computing vector representations of words using dictionary definitions. Dict2Vec also identifies strong pairs of words as words that appear in each other's dictionary definition -- for example, ``car'' might appear in the definition of ``vehicle'' and vice versa, making ``car''/``vehicle'' a strong pair. Synonymy, hypernymy, hyponymy and meronymy are all relationships captured in dictionary definitions and relationships that contribute to high quality clues, so incorporating Dict2Vec into our scoring function allowed us to more heavily weight clues that are semantically related in ways that context alone cannot capture.

We define $DICT(w_1,w_2)$ as the cosine distance between the Dict2Vec word embeddings for two words $w_{1}$ and $w_{2}$. 

\textit{The DETECT score}. Both term relevance $FREQ(w)$ and dictionary relevance $DICT(w_1, w_2)$ are incorporated into a new weighting term, DETECT:

\begin{equation}
\begin{aligned}
DETECT(\tilde{c}) = \lambda_F FREQ(\tilde{c}) + \lambda_D \bigg(\sum_{b \in I}1 - DICT(\tilde{c}, b) - \max_{r \in R}\big{(}1 - DICT(\tilde{c}, r) \big{)} \bigg).
\end{aligned}
\label{eqn:detect}
\end{equation}
A candidate clue $\tilde{c}$ will have a high value for $DETECT(\tilde{c})$ if it is a more common word (without being so common that it falls above $\alpha$ in \textbf{Equation \ref{eqn:idf}}), and is close to as many of the intended blue words as possible in the Dict2Vec embedding space (therefore making $\sum_{b \in I}1 - DICT(\tilde{c}, b)$ as large as possible) while remaining as far as possible from red words. $DETECT(\tilde{c})$ is added to the scoring function $g(\cdot)$ of \textbf{Equation \ref{eqn:our_scoring_fxn}} and $g_{kim}(\cdot)$ of \textbf{Equation \ref{eqn:kim_scoring_fxn}} to re-weight candidate clues. We found $\lambda_F = 2$, $\lambda_D = 1$ for GloVe filtered on the top 10k English words, and $\lambda_D = 2$ for all other representations, to be most effective empirically.

The word embeddings (word2vec, GloVe, fastText, BERT) were obtained from publicly available collections of pre-trained vectors based on large corpora such as Wikipedia or Google News. The word2vec, GloVe, and fastText vectors were obtained from the gensim library \cite{rehurek_lrec}, and BERT contextualized embeddings were obtained from a pre-trained BERT model (bert\_12\_768\_12, book\_corpus\_wiki\_en\_uncased) made available in the GluonNLP package \shortcite{gluoncvnlp2020}. DETECT leverages additional data sources summarized in different ways to improve clue choosing. The Dict2Vec component of DETECT uses dictionary definitions from Cambridge, Oxford Collins, and dictionary.com, and an embedding method to summarize this data. The $FREQ(w)$ component of DETECT uses a cleaned subset of Wikipedia \cite{Mahoney} and the $FREQ(w)$ score to summarize this data. 

\subsection{Using Contextual Embeddings} \label{sec:bertembeddings}

This section describes our final methodological contribution to improve Codenames clue-giving: how to produce a clue using a contextual embedding method. In ``classical'' word embedding methods such as GloVe \cite{pennington2014glove} and word2vec \cite{mikolov2013word2vec}, each word is associated with only one vector representation, so the Codenames scoring function in \textbf{Equation \ref{eqn:our_scoring_fxn}} can be computed directly. However, contextual embedding methods like BERT \cite{devlin-etal-2019-bert} only produce a representation for a word in context. Therefore the word ``running'' in the following sentences, ``She was running to catch the bus,'' and ``She was running for president,'' are captured in their respective contexts. In order to compute the Codenames scoring function using BERT embeddings, we averaged over different contexts to produce a single embedding for each word. Specifically, we extracted contextualized embeddings from a pre-trained BERT model (bert\_12\_768\_12, book\_corpus\_wiki\_en\_uncased), made available by the GluonNLP package \cite{gluoncvnlp2020}, using a cleaned subset of English Wikipedia 2006 \cite{Mahoney}.

Then we defined a word's BERT embedding as the average over that word's contextual embeddings. An approximate nearest neighbor graph was produced from the final embeddings using the Annoy library.\footnote{https://github.com/spotify/annoy} Averaging allowed us to reduce noise caused by outlier contextual embeddings of a word, but in the future we could also experiment with clustering of contextual embeddings and using those clusters to construct a nearest neighbor graph.

\subsection{Human Evaluation of Clue-Giving Performance} \label{sec:humaneval}

We use Amazon Mechanical Turk (AMT) for human evaluation of Codenames clue-giving algorithms.

\subsubsection{General performance comparison and efficacy of DETECT}

We compare five Codenames clue-giving algorithms, based on our proposed scoring function $g(\cdot)$ (\textbf{Equation \ref{eqn:our_scoring_fxn}}): (1) BabelNet-WSF, (2) word2vec \cite{mikolov2013word2vec}, (3) GloVe \cite{pennington2014glove}, (4) GloVe-10k (filtering for the 10k most common words), (5) fastText \shortcite{bojanowski2017enriching}, and (6) BERT \cite{devlin-etal-2019-bert}. \textbf{Table \ref{table:baseline_methods}} summarizes all methods we used in our experiments, including baseline methods from \citeA{kim2019cooperation}, along with all of our new proposed methods and variations on those methods. The symbol ``$\checkmark$'' indicates something novel to the paper, whether it is a new use of a method (first column), a new representation of words (second column), or uses our new ranking method for clues (third column). This table includes the 3 baselines used in \citeA{kim2019cooperation}, as well as 9 new methods proposed in the paper. In our experiments, we evaluate each algorithm with and without DETECT (\textbf{Section \ref{sec:DETECT}}), and with our scoring function and the \citeA{kim2019cooperation} scoring function, for a total of 24 configurations. BabelNet-WSF includes the innovations described in \textbf{Section \ref{sec:babelnetgeneral}}. BERT includes the innovations of \textbf{Section \ref{sec:bertembeddings}}.

\begin{table}[ht]
\centering
 \begin{tabular}{ m{9em}  m{6.3em} m{6.3em} m{6.5em} m{5.5em}  } 
 \toprule
 Word           & new use &  new use  & new knowledge & new ranking\\
 Representation & in Codenames  &  in Codenames & graph method & method\\
  & $g_{kim}(\cdot)$ &  $g(\cdot)$ & in Codenames &\\
 \midrule
BabelNet-WSF & $\checkmark$ & $\checkmark$ & $\checkmark$ & no\\
BabelNet-WSF+DETECT  & $\checkmark$ & $\checkmark$ & $\checkmark$ & $\checkmark$ \\
\hline
BERT & $\checkmark$ & $\checkmark$ & no & no\\
BERT+DETECT  & $\checkmark$ & $\checkmark$ & no & $\checkmark$\\ 
\hline 
fastText & $\checkmark$ & $\checkmark$ & no & no\\
fastText+DETECT  & $\checkmark$ & $\checkmark$ & no & $\checkmark$ \\
\hline
GloVe & $\checkmark$ & $\checkmark$ & no & no\\
GloVe+DETECT  & $\checkmark$ & $\checkmark$ & no & $\checkmark$ \\
\hline
GloVe-10k& no & $\checkmark$ & no & no\\
GloVe-10k+DETECT  & $\checkmark$ & $\checkmark$ & no & $\checkmark$ \\
\hline
word2vec & $\checkmark$ & $\checkmark$ & no & no\\
word2vec+DETECT & $\checkmark$ & $\checkmark$ & no & $\checkmark$\\
\bottomrule
\end{tabular}
\caption{A summary of all methods from our experiments, both baseline methods from \citeA{kim2019cooperation} as well as new proposed methods and the variations on those methods. The first column indicates whether an algorithm is a new method for Codenames using $g_{kim}(\cdot)$ scoring function, the second column indicates whether it is a new method for Codenames using $g(\cdot)$ scoring function, the third column indicates whether a new knowledge graph method is used, and the fourth column indicates the use of a new method for ranking clues.}
\label{table:baseline_methods}
\end{table}

To compare the algorithms, 60 unique Codenames boards of 20 words each were computationally generated from a list of 208 words obtained from the official Codenames cards. We used words from the official Codenames cards because these words were carefully selected by the game designers to have interesting, inter-related multiple meanings, and as such these words are an integral part of the game definition. 

For a given Codenames board, each algorithm was asked to output the best clue and the two words intended to match the clue. These two words are the blue words that the algorithm based a particular clue on, which a human guesser is intended to select.

For human evaluation, United-States-based AMT workers with a high approval rate ($\geq 98\%$) on previous AMT tasks were asked to look at a full board of 20 words and 1 clue word, and rank the top four board words matching the given clue. The user was required to fill in Ranks 1 and 2, because the algorithm always intended 2 words to be selected. Ranks 3 and 4 were optional for the user to fill in; the AMT worker could specify ``no more related words'' for these ranks. Ranks 3 and 4 were included to distinguish between situations where the algorithm produced a wholly irrelevant clue (such that the worker could not guess the intended words even if given 4 slots) and a sub-optimal clue (such that the worker could guess intended words in Ranks 3 or 4, but not Ranks 1 or 2). The AMT workers had no knowledge of the underlying algorithm and no knowledge of the algorithm's ``intended words.'' A high-performing algorithm chooses such good quality clues that the AMT workers correctly guess the intended words and place them in Ranks 1 and 2.

\subsubsection{Comparison with \citeA{kim2019cooperation}}

In order to compare with the work of \citeA{kim2019cooperation}, we also evaluated the performance of each algorithm using their scoring function $g_{kim}(\cdot)$ (\textbf{Equation \ref{eqn:kim_scoring_fxn}}), using the same set of Codenames boards. In the original formulation of the \citeA{kim2019cooperation} scoring function shown in \textbf{Equation \ref{eqn:kim_scoring_fxn}}, the constraints $\min_{b \in \bc}s( \tilde{c} , b) > \lambda_T$ and $\min_{b \in \bc}s( \tilde{c} , b) > \max_{r \in R}s( \tilde{c} , r)$ were used. Since we evaluate performance on the clue-giving task considering $m=2$ for consistency of evaluation across all algorithms, this constraint was relaxed if there were no clue words passing those constraints for $m=2$ blue words.

In addition, \citeA{kim2019cooperation} restricts to the top 10k common English words. We tested this using the GloVe model filtered on the top 10k common words, and refer to this as GloVe-10k in our reported results. 

\subsubsection{Performance metrics}

Each trial from Amazon Mechanical Turk provided the 2 to 4 board words (ranked) that the AMT worker selected as most related to the given clue word. To quantify the performance of different algorithms, we calculated precision@2 and recall@4. Precision@2 measures the number of correct words that an AMT worker guessed in the first 2 ranks, where correct words are the intended words that the algorithm meant the worker to select for a given clue. Recall@4 measures the number of correct words chosen in the first 4 ranks.

\section{Results}

 \begin{figure*}[ht]
    \centering
    \includegraphics[scale=0.3]{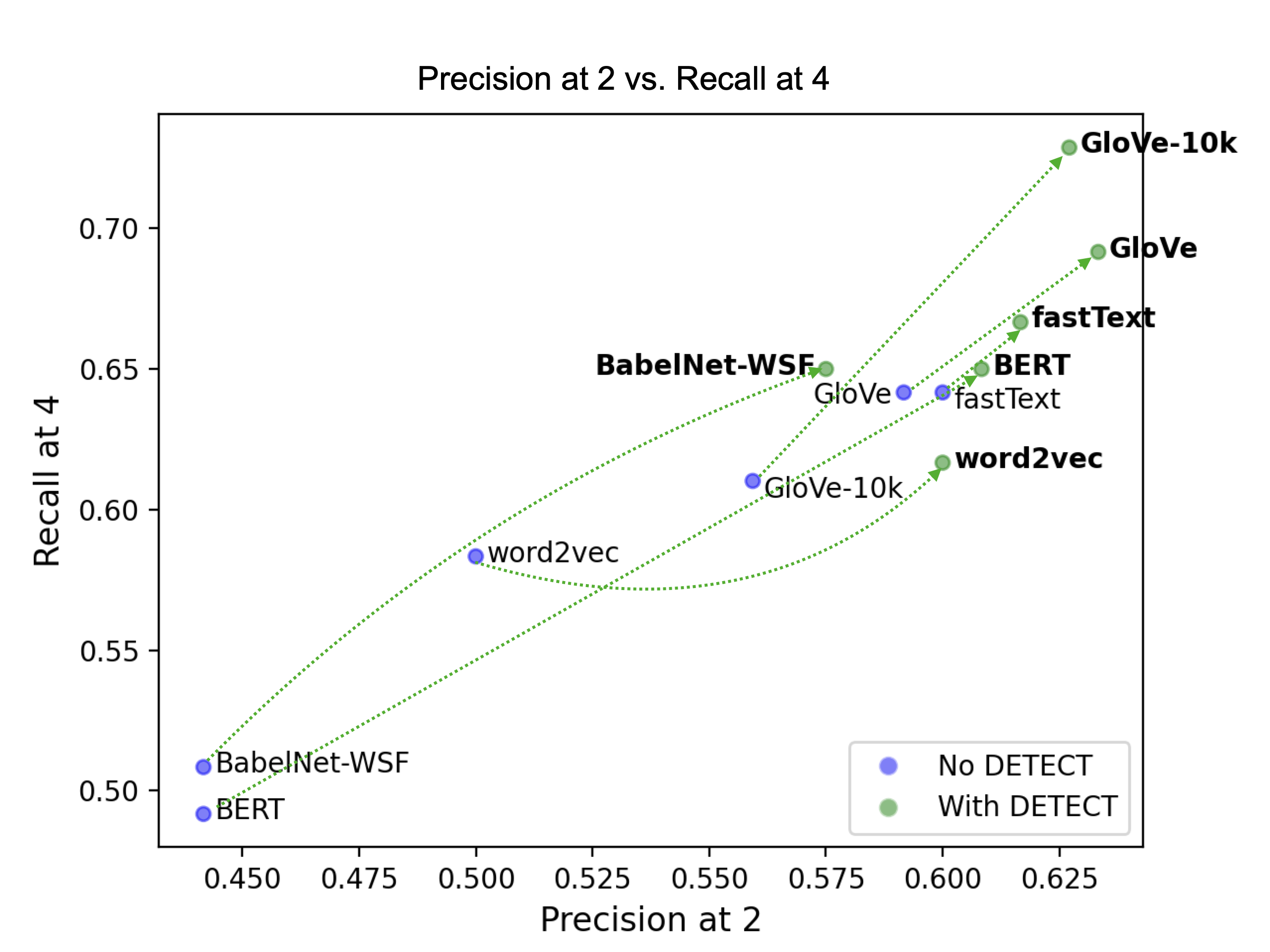}
    \caption{Precision@2 (for the 2 intended words chosen by the algorithm for a clue) vs. Recall@4 (for the 4 words that guessers could answer) from Amazon Mechanical Turk results using our proposed scoring function $g(\cdot)$.}
    \label{fig:our_results}
\end{figure*}

\begin{figure*}[ht]
    \centering
    \includegraphics[scale=0.3]{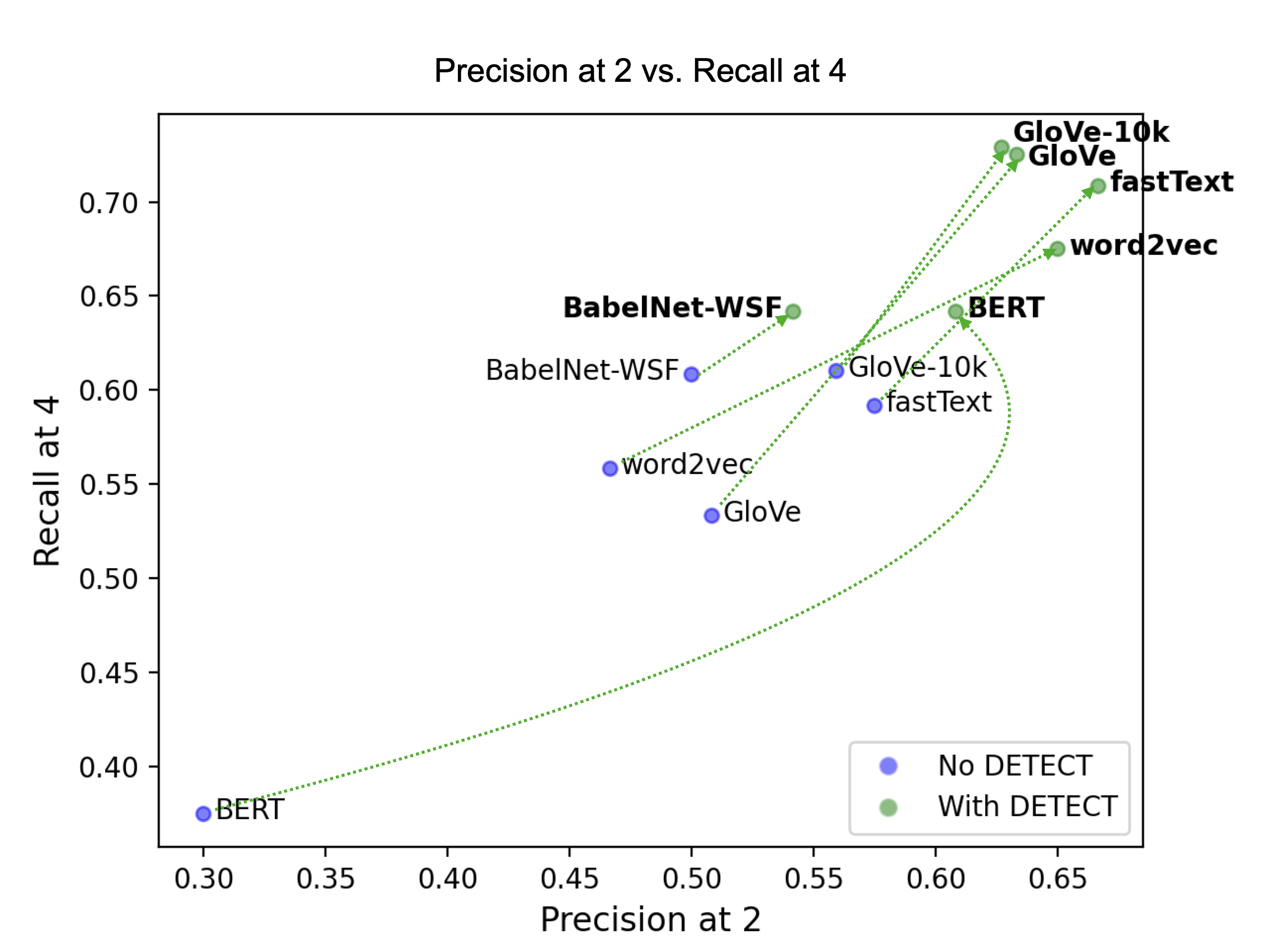}
    \caption{Precision@2 (for the 2 intended words chosen by the algorithm for a clue) vs. Recall@4 (for the 4 words that guessers could answer) from Amazon Mechanical Turk results using the \citeA{kim2019cooperation} scoring function.}
    \label{fig:kim_results}
\end{figure*}

Results from the different algorithms with our scoring function are shown in \textbf{Figure \ref{fig:our_results}}, while results for the different algorithms with the \citeA{kim2019cooperation} scoring function are shown in \textbf{Figure \ref{fig:kim_results}}. Additional results from the AMT evaluation are reported in \textbf{Appendix A}. The results are summarized as follows.

\begin{itemize}
\item Some of our methods surpass state-of-the-art performance for the Codenames clue-giving task. The best-performing algorithm for precision@2 across both scoring functions is fastText+DETECT, with 66.67\% precision@2.
\textbf{fastText+DETECT outperforms the prior state-of-the-art by \citeA{kim2019cooperation} using glove-10k, which has precision 55.93\% as shown in \textbf{Figure \ref{fig:kim_results}}.}

\item \textbf{Our proposed DETECT algorithm leads to universal improvement across all word representations} with a median percent improvement of 18.0\% for precision@2. DETECT also leads to improvement across both scoring functions, with a median 16.1\% improvement for our scoring function and even higher median improvement of 20.3\% for the \citeA{kim2019cooperation} scoring function. For BERT, the advantage of using DETECT is most substantial, with a 102.8\% improvement in  precision@2.

\item When DETECT is used, our scoring function leads to better performance for BabelNet-WSF, while the \citeA{kim2019cooperation} scoring function leads to better performance for word2vec and fastText, and both performed equally on GloVe, GloVe-10k, and BERT (precision@2). Thus, neither scoring function is definitively superior. However, since BabelNet-WSF is more interpretable and easier to troubleshoot than the word embedding methods, we suspect that our scoring function (which performs better on BabelNet-WSF) might be more useful in settings where one might want to engineer better performance for the system.

\item \textbf{We report the first successful use of a knowledge graph to solve the Codenames task}, with 57.5\% precision@2 for BabelNet-WSF with our scoring function, comparable to the performance of word embedding-based methods.
\end{itemize}

\section{Discussion}

Codenames is a task that is difficult even for humans, who sometimes struggle to generate clues for particular boards. As with all games, the element of difficulty is necessary in order to produce an exciting challenge. Codenames relies on a deep understanding of language. Traditional language tasks often focus on one axis of language understanding such as analogies or part-of-speech tagging, while Codenames requires leveraging many different axes of language, including common sense relationships between words. 

In this work, we propose several innovations that improve performance on the Codenames clue-giving task. First, we enable successful clue-giving from a knowledge graph, BabelNet-WSF, via new techniques for sub-graph construction, nearest neighbor filtering, and single word clue selection from multi-word synset phrases. Our innovations enable BabelNet-WSF performance that is comparable to word embedding-based methods, while retaining the advantage of full interpretability due to the underlying graph structure. Next, we propose DETECT, a score that combines document frequency and Dict2Vec embeddings to eliminate too-rare or too-common potential clues while incorporating more diverse word relationships. DETECT improves performance across all algorithms and scoring functions. Finally, we complete the first large-scale human evaluation of Codenames algorithms on Amazon Mechanical Turk, to accurately evaluate the real-world performance of our algorithms. Overall, our proposed methods yield state-of-the-art performance on Codenames and advance the formal study of word games.

\newpage

\appendix

\section{Examples}
\begin{figure}[H]
    \centering
    \includegraphics[scale=0.25]{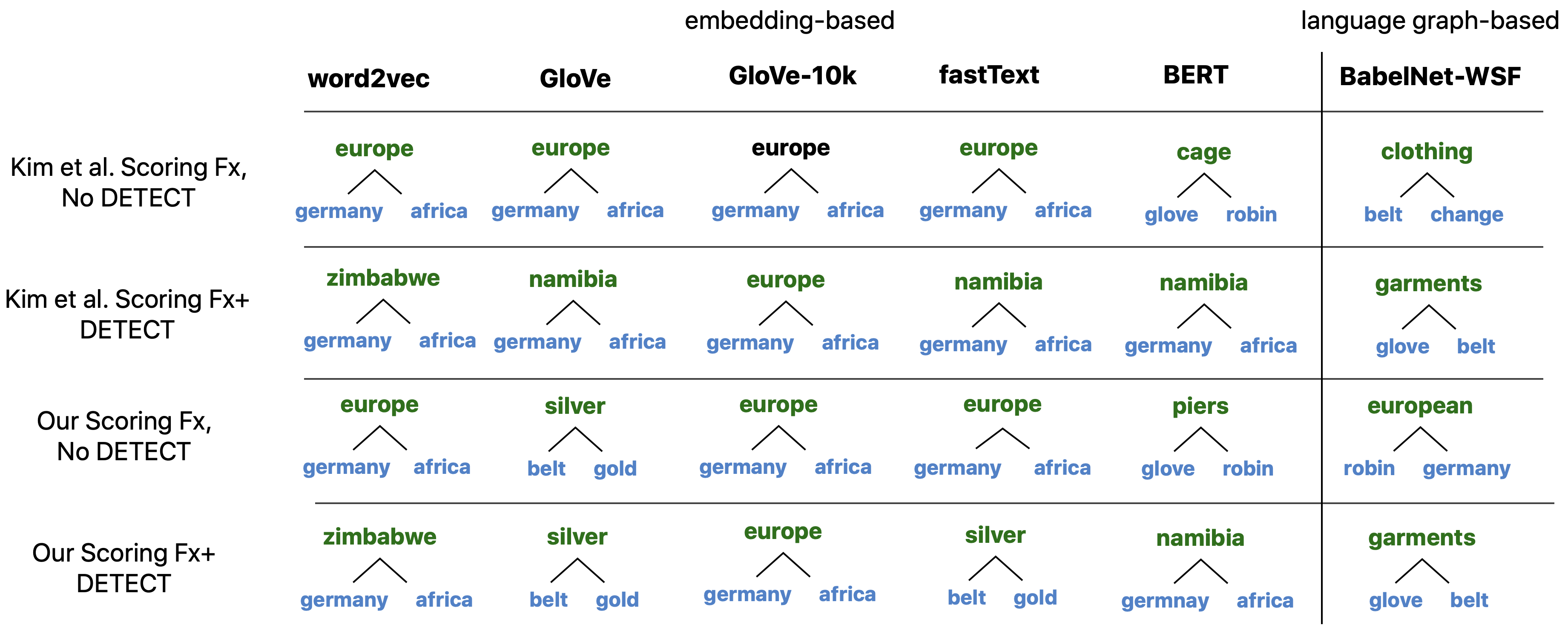}
    \caption{The clues chosen (in black for baselines and green for our methods) and the intended board words chosen for that clue (in blue) for all experimental methods. This includes the 6 word representations (word2Vec, GloVe, GloVe-10k, fastText, BERT, BabelNet-WSF), 2 scoring functions (ours and Kim et al.'s), and with and without DETECT applied. The board words for this trial were blue = germany, car, change glove, needle, robin, belt, board, africa, gold; red = pipe, kid, key, boom, satellite, tap, nurse, pyramid, rock, bark.}
    \label{fig:large_examples1}
\end{figure}

\begin{figure}[H]
    \centering
    \includegraphics[scale=0.25]{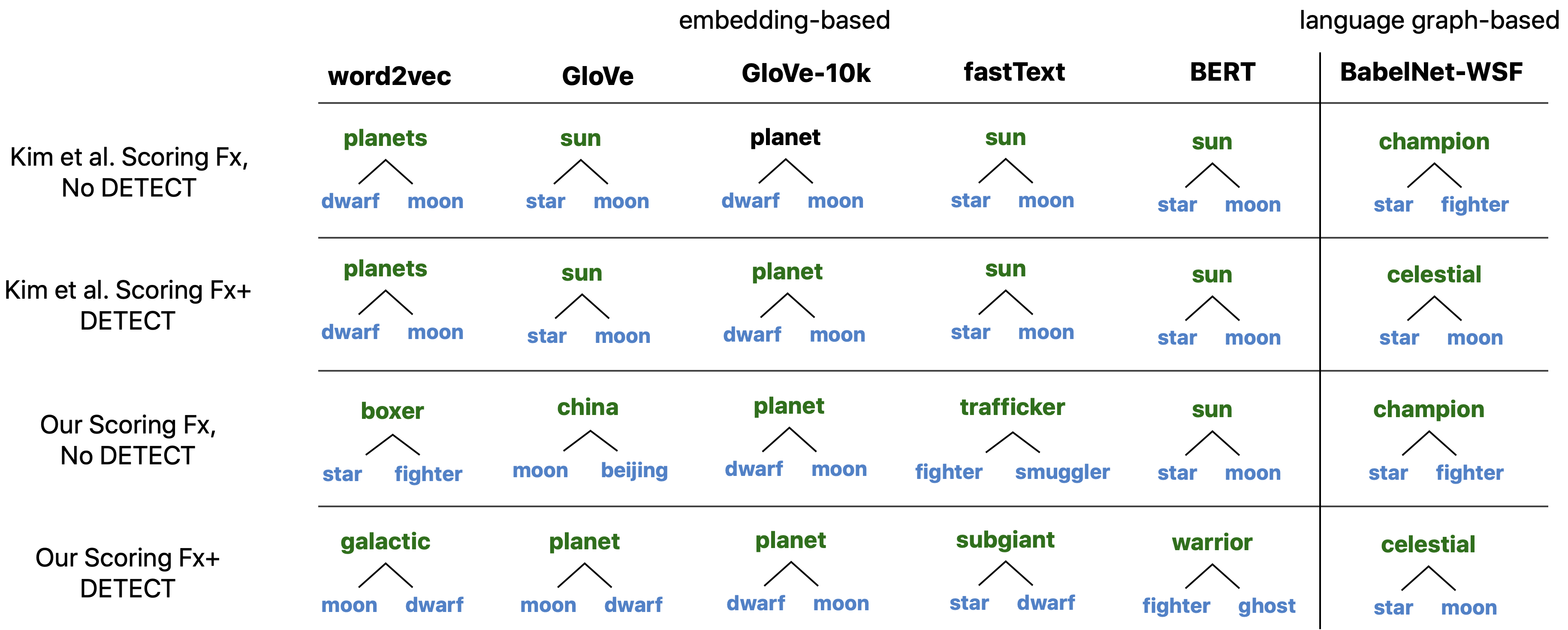}
    \caption{The clues chosen (in black for baselines and green for our methods) and the intended board words chosen for that clue (in blue) for all experimental methods. The board words for this trial were blue = dwarf, foot, moon, star, ghost, beijing, fighter, roulette, alps; red = club, superhero, mount, bomb, knife, belt, robot, rock, bar, lab.}
    \label{fig:large_examples2}
\end{figure}

\begin{figure}[H]
    \centering
    \includegraphics[scale=0.25]{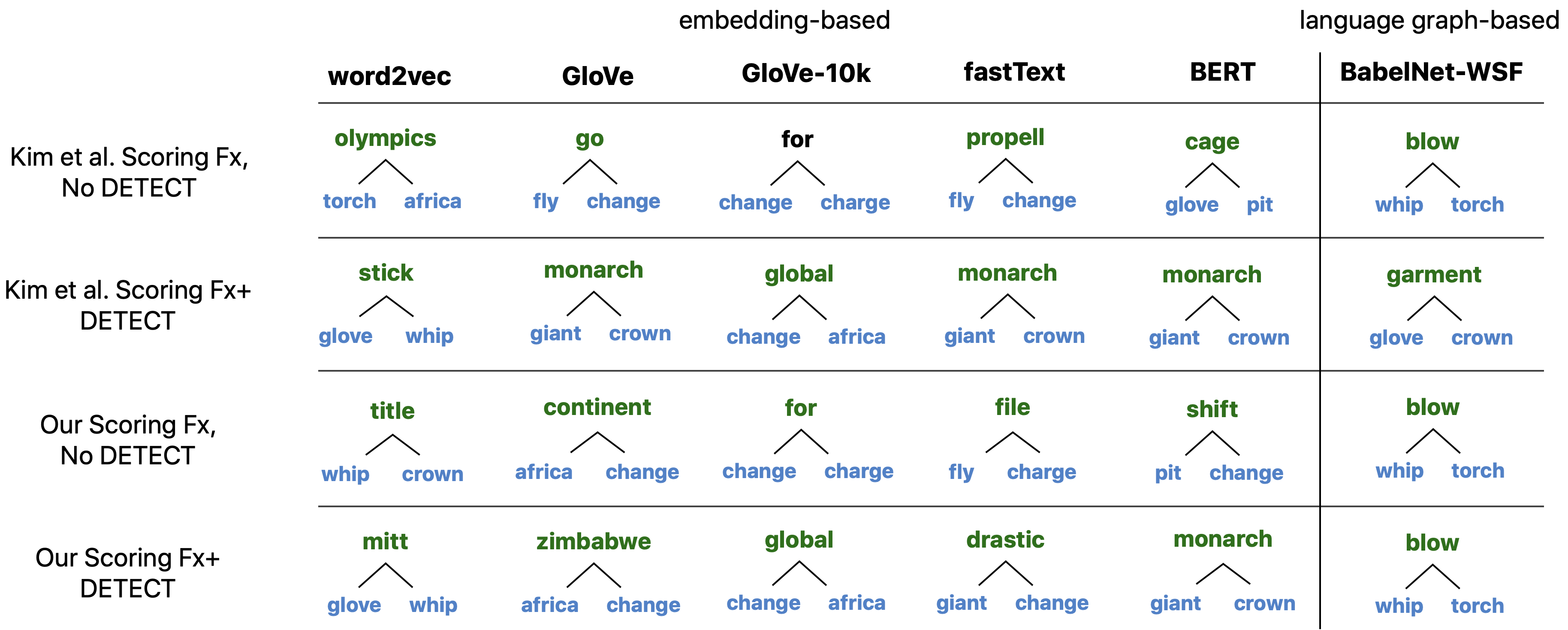}
    \caption{The clues chosen (in black for baselines and green for our methods) and the intended board words chosen for that clue (in blue) for all experimental methods. The board words for this trial were blue = crown, pit, change, glove, charge, torch, whip, fly, africa, giant; red = amazon, hole, shark, ground, shop, cast, nurse, server, vacuum, rock.}
    \label{fig:large_examples3}
\end{figure}

\begin{figure}[H]
    \centering
    \includegraphics[scale=0.25]{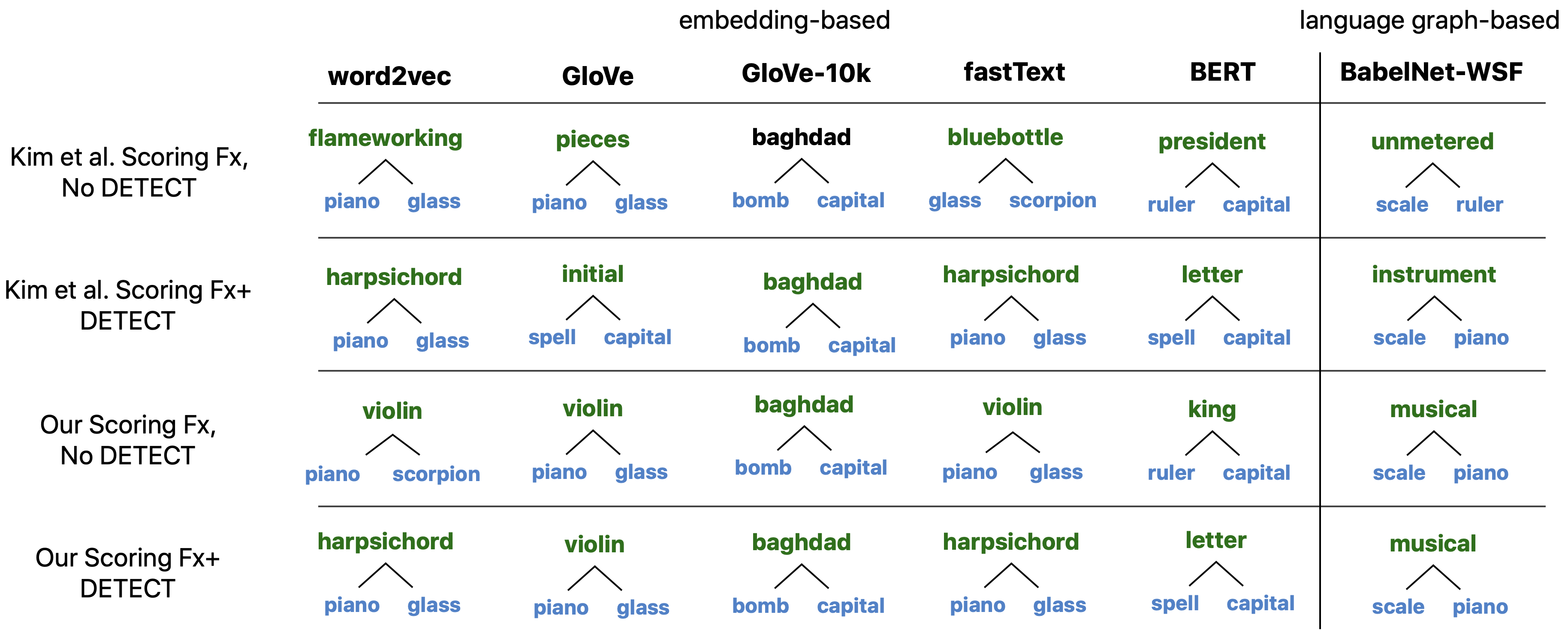}
    \caption{The clues chosen (in black for baselines and green for our methods) and the intended board words chosen for that clue (in blue) for all experimental methods. The board words for this trial were blue = amazon, spell, ruler, scale, round, bomb, piano, glass, capital, scorpion; red = paste, air, ground, cold, lemon, belt, torch, point, saturn, game.}
    \label{fig:large_examples4}
\end{figure}

\textbf{Figures \ref{fig:large_examples1}-\ref{fig:large_examples4}}  show examples of clues chosen by all the experimental configuration for four Codenames boards.

\section{Algorithms}
\textbf{Algorithm \ref{algo:2}} details the algorithm for getting single word clues from BabelNet, as described in \textbf{Section \ref{sec:single_word_clues}}. \textbf{Algorithm \ref{algo:3}} details how nearest neighbors are queried for and cached from BabelNet, as described in \textbf{Section \ref{sec:constructing_subgraph}}.

\begin{algorithm}[t]
\SetAlgoLined
\SetKwInOut{Input}{Input}\SetKwInOut{Output}{Output}
\Input{\mainSense, a string representing the main sense label of a synset. \otherSenses, a list of strings representing the other sense labels of the synset. In both \mainSense{} and \otherSenses, multi-word clues delimited by the `\_' character. $w_1$, $w_2$, $w_3$, $w_4$, corresponding to weights described in \textbf{Table \ref{table:1}}}
\Output{a dictionary of single word labels and scores corresponding to the configured weights for each label type}
\BlankLine
\Begin{
    \singleWordLabels $\longleftarrow \emptyset$\ \Comment*[r]{Initialize \singleWordLabels as an empty set}
    \splitMainSense = \textsc{split}(\mainSense,`$\_$')\ \Comment*[r]{Split main sense label using delimiter~$\_$}
    \If(\Comment*[f]{Main sense label is a single word}){\len(\splitMainSense) = 1} { 
        \singleWordLabels[\splitMainSense[0]] = $w_1$\;
    }
    \Else(\Comment*[f]{Main sense label is multiple words})
    { 
    \For(\Comment*[f]{Set each word's weight to $w_2$}){\word{} $\in$ \splitMainSense}
        { 
            \singleWordLabels[\word] = $w_2$\; 
        }
     }
    \For(\Comment*[f]{Iterate over other sense labels}){\sense{} $\in$ \otherSenses} { 
        \splitOtherSense = \textsc{split}(\sense,`$\_$')\ \Comment*[r]{Split other sense label using delimiter~$\_$}
        \If(\Comment*[f]{Other sense label is a single word}){\len(\splitOtherSense) = 1} 
        { 
            \singleWordLabels[\splitOtherSense[0]] = $w_3$\;
        }
        \Else(\Comment*[f]{Other sense label is multiple words})
        { 
            \For(\Comment*[f]{Set each word's weight to $w_4$}){\word{} $\in$ \splitOtherSense} 
            {
                \singleWordLabels[\word] = $w_4$\;
            }
        }
    }
    return \singleWordLabels\;
}
\caption{Extracting single-word clues for a synset\label{algo:2}}
\end{algorithm}

\begin{algorithm}[t]
\SetAlgoLined
\SetKwInOut{Input}{Input}\SetKwInOut{Output}{Output}
\Input{\lemmaSynsets, a dictionary of synsets for each board word. $B$ and $R$, the set of our team's and the other team's board words, respectively. $L$, number of levels (the number of edges from source word) to query}
\Output{a dictionary of board words and synset edges at each level of edges to be cached}
\BlankLine
\Begin{
    \For(\Comment*[f]{Iterate over the union of blue and red teams' board words}){\word{} $\in B \cup R$}{ 
        \synsets{} $\longleftarrow$ \lemmaSynsets[\word]\ \Comment*[r]{Get the synsets mapped to board word}
        \For(\Comment*[f]{Repeat for each \level{} from $1$ to $L$, where \level{} is the number of edges from the board word}){\level $\longleftarrow$ 1 to $L$} { 
            \nextLevelSynsets{} $\longleftarrow \emptyset$\ \Comment*[r]{Initialize \nextLevelSynsets{} to empty set}
            \For{\synset{} $\in$ \synsets}{ 
                \edges{} $\longleftarrow$ \textsc{getOutgoingEdges}(\synset)\ \Comment*[r]{Get the outgoing edges from this synset in the BabelNet graph}
                $D$[\word][\synset][\level] $\longleftarrow$ \edges\ \Comment*[r]{Store the edges mapped to this board word, synset, and level}
                \For{\edge{} $\in$ \edges} { 
                    \If{\textsc{!isAutomatic}(\edge) \textbf{\upshape and} \textsc{relationGroup}(\edge) = \textsc{hypernym}} {
                        \nextLevelSynsets{} $\longleftarrow$ \nextLevelSynsets + \textsc{getConnectedSynset}(\edge)\ \Comment*[r]{If the edge is non-automatic and in relation group \textsc{hypernym}, add the connected synset to the set of synsets to query for the next level}
                    } 
                }
            }
            $synsets \longleftarrow nextLevelSynsets$\;
        }
    }
    return $D$\;
}
\caption{Querying BabelNet Edges\label{algo:3}}
\end{algorithm}

\clearpage
\section{Amazon Mechanical Turk Results}
\textbf{Table \ref{table:amt_results}} shows Amazon Mechanical Turk results for 1,400 trials, using our scoring function and \citeA{kim2019cooperation}'s scoring function, as well as with and without applying DETECT to the clue selection algorithm.

\begin{table}[ht]
\centering
 \begin{tabular}{ m{12em}  m{7em}  m{5em}  m{7em}  m{5em} } 
 \toprule
\multicolumn{1}{c}{} &  \multicolumn{2}{c}{\textbf{$g(\cdot)$}} & \multicolumn{2}{c}{\textbf{$g_{kim}(\cdot)$}}\\
 Word Representation & Precision@2 & Recall@4 & Precision@2 & Recall@4 \\
 \midrule
BabelNet-WSF               & 0.442    & 0.508   & 0.500 & 0.608   \\
\textbf{BabelNet-WSF+DETECT} & \textbf{0.575*} &\textbf{0.650*} & 0.542 & 0.642  \\
\hline
BERT                   & 0.442   & 0.492      & 0.3  & 0.375  \\
\textbf{BERT+DETECT}   & \textbf{0.608*} & \textbf{0.65*}  & \textbf{0.608*} & \textbf{0.642*}\\ 
\hline 
fastText               & 0.6 & 0.642  & 0.575 & 0.592\\
\textbf{fastText+DETECT} & 0.617  & 0.667  & 0.667 & 0.708  \\
\hline
GloVe                  & 0.592    & 0.642  & 0.508  & 0.533  \\
\textbf{GloVe+DETECT}    & 0.633 & 0.692 & \textbf{0.633*} & \textbf{0.725*} \\
\hline
GloVe-10k                & 0.560    & 0.61  & 0.56    & 0.61   \\
\textbf{GloVe-10k+DETECT}  & 0.627 & \textbf{0.729*}    & 0.627 & \textbf{0.729*}          \\
\hline
word2vec               & 0.5 & 0.583 & 0.467  & 0.558\\
\textbf{word2vec+DETECT} & 0.6 & 0.617  & \textbf{0.65*}  & 0.675 \\

\bottomrule
\end{tabular}

\caption{Amazon Mechanical Turk results for our proposed scoring function $g(\cdot)$  and \citeA{kim2019cooperation} scoring function $g_{kim}(\cdot)$ for intended word precision@2 (number of intended words chosen in the first 2 ranks/2) and  intended word recall@4 (number of intended words chosen in the first 4 ranks/2). * indicates statistical significance ($p < 0.05$) between the word representation and word representation +DETECT  using a z-test. }
\label{table:amt_results}
\end{table}

\clearpage
\bibliographystyle{theapa}
\bibliography{sample}

\end{document}